\documentclass[letterpaper, 10 pt, conference]{ieeeconf}  
\pdfoutput=1

\newif\ifarxiv
\arxivfalse

\newif\ifcaptionmod
\captionmodfalse
\IEEEoverridecommandlockouts                              

\usepackage{graphics,graphicx} 
\graphicspath{{../Figures/}}
\usepackage{amsmath} 
\usepackage{amssymb}  
\usepackage{xcolor}

\usepackage{amsmath} 
\usepackage{amssymb}  
\usepackage{xcolor}
\usepackage{algorithm}
\usepackage{algpseudocode}
\usepackage{cite}
\usepackage{dblfloatfix}
\usepackage{mathtools} 
\usepackage{dutchcal} 
\usepackage{caption}
\usepackage{subcaption}
\usepackage{hyperref}
\usepackage{lipsum}

\newcommand{\norm}[1]{\left\lVert#1\right\lVert}
\newcommand{\snorm}[1]{\norm{#1}^2}
\newcommand{\abs}[1]{\left\lvert#1\right\rvert}

\newcommand{\x}{\mathbf{x}}

\newcommand{\f}{\mathbf{f}}

\newcommand{\q}{\mathbf{q}}

\newcommand{\g}{\mathbf{g}}

\newcommand{\blambda}{\boldsymbol{\lambda}}

\newcommand{\p}{\mathbf{p}}
\newcommand{\T}{\mathbf{T}}
\newcommand{\Q}{\mathbf{Q}}
\newcommand{\R}{\mathbf{R}}

\newcommand{\J}{\mathbf{J}}

\newcommand{\I}{\mathbf{I}}
\newcommand{\h}{\mathbf{h}}
\newcommand{\bomega}{\boldsymbol{\omega}}
\newcommand{\btheta}{\boldsymbol{\theta}}

\renewcommand{\v}{\mathbf{v}}
\renewcommand{\S}{\mathbf{S}}

\renewcommand{\u}{\mathbf{u}}
\renewcommand{\P}{\mathbf{P}}
\renewcommand{\S}{\mathbf{S}}
\renewcommand{\u}{\mathbf{u}}

\renewcommand{\xi}{\x^{[i]}}
\newcommand{\ui}{\u^{[i]}}
\newcommand{\gi}{\g^{[i]}}
\newcommand{\hi}{\h^{[i]}}

\title{\LARGE \bf
Zero-Shot Retargeting of Learned Quadruped Locomotion Policies Using Hybrid Kinodynamic Model Predictive Control
}

\author{He Li$^{1}$, Wenhao Yu$^{2}$, Tingnan Zhang$^{2}$, Patrick M. Wensing$^{1}$ 
\thanks{*This work was supported by a research gift from Google LLC to the University of Notre Dame}
\thanks{$^{1}$ University of Notre Dame, Notre Dame, IN, USA}%
\thanks{$^{2}$ Robotics at Google, Mountain View, CA, USA}
}

\newcommand{\HL}[1]{\textcolor{black}{#1}}

\begin{document}

\maketitle
\thispagestyle{empty}
\pagestyle{empty}

\begin{abstract}
Reinforcement Learning (RL) has witnessed great strides for quadruped locomotion, with continued progress in the reliable sim-to-real transfer of policies. However, it remains a challenge to reuse a policy on another robot, which could save time for retraining. In this work, we present a framework for zero-shot policy retargeting wherein diverse motor skills can be transferred between robots of different shapes and sizes. The new framework centers on  a planning-and-control pipeline that systematically integrates RL and Model Predictive Control (MPC). The planning stage employs RL to generate a dynamically plausible trajectory as well as the contact schedule, avoiding the combinatorial complexity of contact sequence optimization. This information is then used to seed the MPC to stabilize and robustify the policy roll-out via a new Hybrid Kinodynamic (HKD) model that implicitly optimizes the foothold locations. Hardware results show an ability to transfer policies from both the A1 and Laikago robots to the MIT Mini Cheetah robot without requiring any policy re-tuning. 
\end{abstract}


\section{Introduction}
Over the past decade, the promise of legged robots benefiting our daily life has dramatically increased following successful demonstrations from several legged robot platforms such as Atlas, Spot, ANYmal \cite{hutter2016anymal}, MIT Cheetah 3 \cite{bledt2018cheetah}, and many others. Whole-body dynamic controllers have been developed to unlock the motions of these platforms, with many based on two techniques, Model Predictive Control (MPC) and Reinforcement Learning (RL).

MPC methods can stabilize legged locomotion by generating control signals via Trajectory Optimization (TO) that respects various constraints. Successful implementations of MPC have been demonstrated on many of the platforms mentioned above \cite{di2018dynamic, kuindersma2016optimization, bellicoso2018dynamic}. Since the robot dynamics and constraints are most often nonlinear, solving such TO problems could be time-consuming and  prone to local optima. Despite these challenges, a few notable previous works have shown success of whole-body MPC \cite{farshidian2017real, neunert2016fast} on quadruped robots. A common approach to mitigate computational barriers is to simplify the dynamics, for example, by adopting a kinodynamic model \cite{grandia2019feedback} that considers the body dynamics and leg kinematics of a quadruped. Even simpler, the convex MPC \cite{di2018dynamic} linearizes body dynamics around a reference trajectory and ignores the legs during planning. Nevertheless, it remains nontrivial to obtain a good initial guess for whole-body MPC or a good reference for convex MPC for synthesizing and controlling complex motions.

RL does not suffer from some of these problems and can learn control policies for general tasks \cite{peng2018deepmimic, peng2020learning, hwangbo2019learning, lee2020learning, miki2022learning}. These benefits are realized by extensively exploring the robot dynamics in a Monte Carlo manner. The extensive search in robot control and state space enables the RL policy to discover complex motions. Peng et al.~\cite{peng2018deepmimic} developed DeepMimic using RL for humanoid robots to imitate reference motion capture segments, with which, the simulated Atlas robot could perform a spinkick. In a separate work, they also developed an imitation RL policy for quadruped robots to imitate the motions of dogs, which was verified on hardware as well. More appealing results are revealed in \cite{miki2022learning} where the ANYmal robot could traverse very challenging terrains with RL. In general, nevertheless, RL remains subject to sim-to-real transfer difficulty due to the inaccurate modeling of the robot dynamics and demands sim-to-real transfer techniques such as dynamics randomization \cite{peng2018sim, hwangbo2019learning,peng2020learning,miki2022learning}, which often increases training difficulties. Further, training an RL policy could be time-consuming in terms of building the training environment, tuning hyper parameters, and through the training process itself. 

\begin{figure}[t]
    \centering
    \includegraphics[width = 0.9\linewidth]{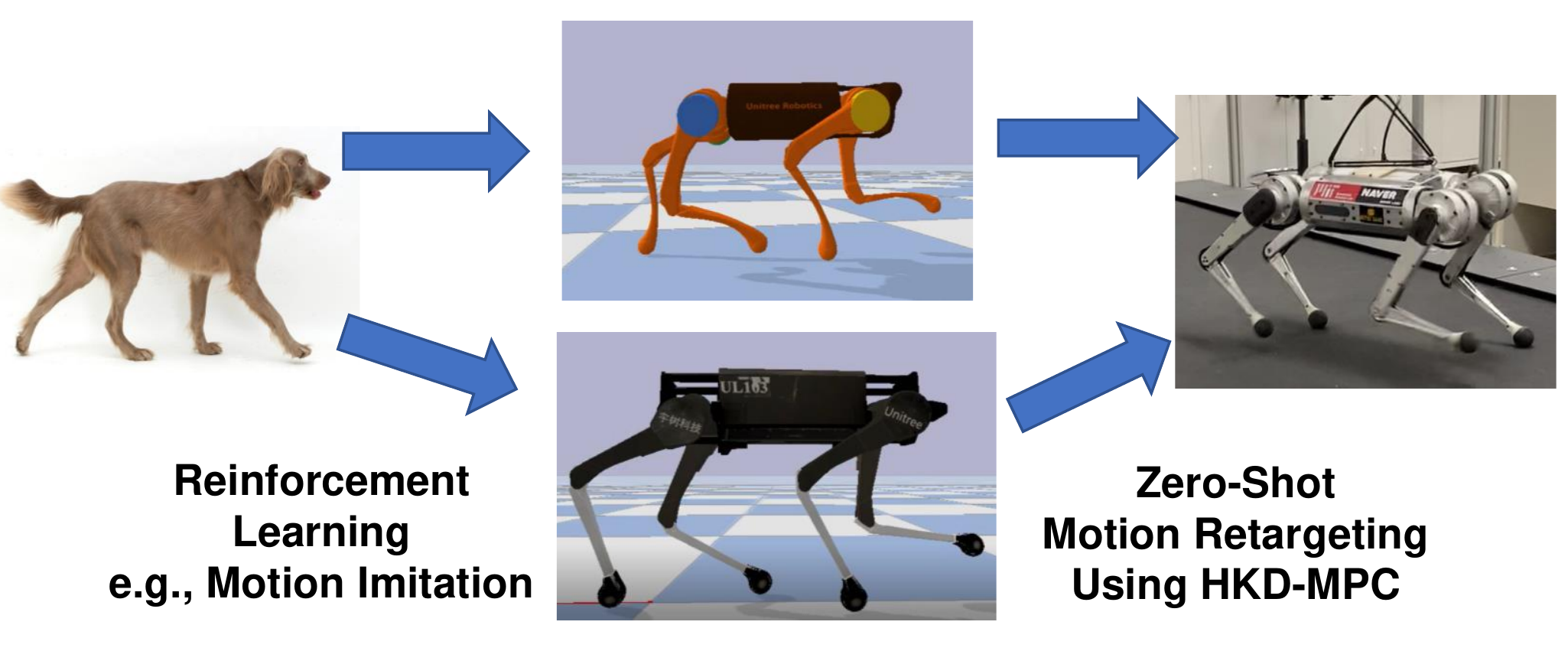}
    \caption{Summary of results: RL policies are trained to imitate dog motions on Unitree A1 (middle top) and Laikago (middle bottom). Both policies are retargeted to MIT Mini Cheetah hardware without further tuning.}
    \label{fig:deliverable}
\end{figure}

Suppose now an RL policy is ready to use with one robot, and where we would like to achieve similar behaviors on another robot with  the same topology but different  geometric and physical parameters. 
An immediate question concerns whether information encoded in the existing policy can be effectively transferred to the new system. Direct deployment of the RL policy is likely not possible since it is sensitive to the model accuracy, which also accounts for the well-known sim-to-real transfer problem.

In this work, we propose a novel perspective on retargeting the RL policy to different robot platforms by integrating RL and MPC for mutual benefits.
In specific, we obtain a trajectory reference and contact schedule by unrolling the RL policy, which then seeds an MPC solve for a new robot. The MPC solver then synthesizes a motion that is dynamically feasible for the target robot platform, while trying to follow the RL roll-out to the maximal extent. 
This approach enables us to easily leverage the vast body of existing work and pre-trained models in DRL to acquire complex motor skills for our robot \cite{peng2018deepmimic,peng2020learning}.
In this work, we use the motion imitation learning framework developed in \cite{peng2020learning} for RL policy training. However, the proposed retargeting framework may potentially be used with other RL policies in general. A summary of the results of this work is shown in Fig.~\ref{fig:deliverable}. The RL policies learn trotting and pacing on two quadruped robots, Unitree A1 and Laikago, and are then both successfully retargeted to the MIT Mini Cheetah without additional training.

\subsection{Related Works}
There are many existing previous works that have investigated combing learning and optimization-based control for quadruped locomotion. GLIDE \cite{xie2021glide} employs RL to predict the CoM acceleration at the next time step and uses a centroidal-model QP for producing the ground reaction force and maintaining balance. RLOC \cite{gangapurwala2020rloc} develops perception-aware RL for planning the centroidal trajectory at the next time step, as well as the foot placement, and employs a whole-body controller for the control. They demonstrate the transferability of a pipeline originally trained for ANYmal B to the larger and heavier ANYmal C. This transfer is accomplished by training a domain adaptive tracker that produces corrective joint torques based on a history of tracking errors. It is not uncommon to imitate animal motions on quadruped robots. One seminal work toward this direction is by Peng et al.~\cite{ peng2020learning} where they use RL to imitate dog motions on a quadruped robot. Other works such as \cite{li2021model} use model based methods. They use Dynamic Movement Primitives (DMP) to fit the reference animal motions by solving TO problems in a loop, demonstrating the need to use a dynamically-feasible trajectory as a reference in place of the original animal motion.

The contributions of this work are two-fold. (1) A systematic control architecture that enables zero-shot retargeting of motor skills on robots with different geometric and physical parameters. (2) An MPC controller based on a hybrid kinodynamic (HKD) model that can simultaneously optimize the foothold locations, the Center of Mass (CoM) trajectory, and ground reaction forces (GRF). Compared to the kinodynamic model \cite{grandia2019feedback}, the HKD model only considers the leg kinematics in swing, and uses the foothold location at touchdown for the stance phase, thus avoiding the need to explicitly enforce a non-sliding constraint during stance. We demonstrate that with the proposed MPC controller, policies could be retargeted to the MIT Mini Cheetah after being originally trained for two other quadruped robots (A1 and Laikago) of different size and weight.

\begin{figure*}[!t]
    \centering
    \includegraphics[width = .8\linewidth]{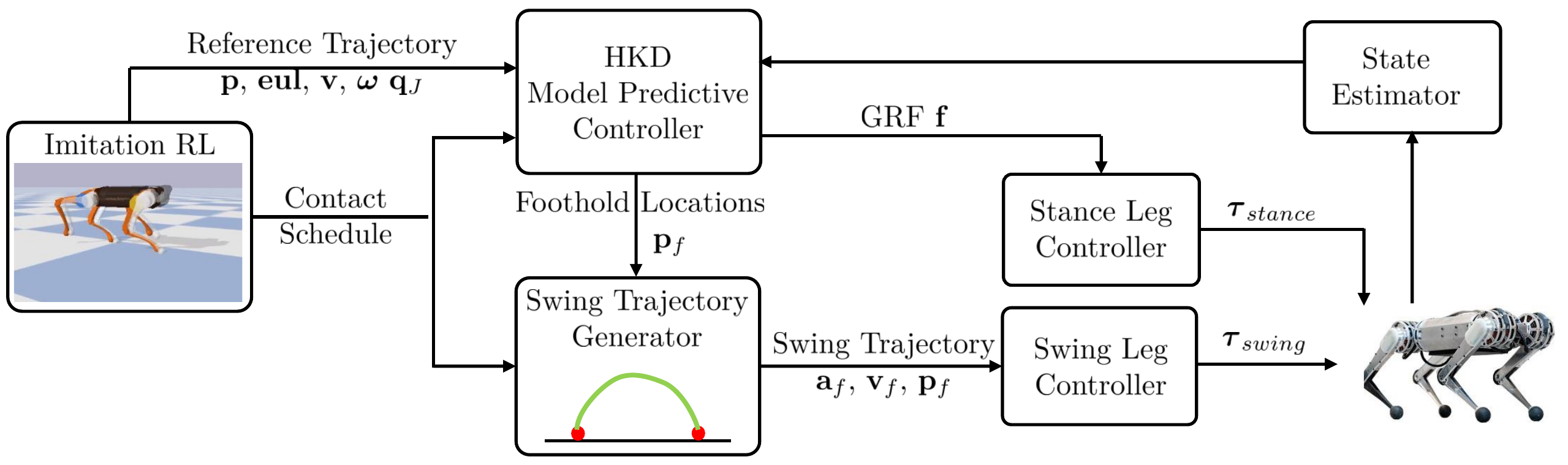}
    \caption{Overview of the proposed control architecture for motion retargeting.}
    \label{fig:overview}
\end{figure*}

\section{Background}
\subsection{Reinforcement Learning}
In this section, we introduce basic concepts in RL and encourage the readers to refer to \cite{RLIntro} for details. The goal of RL is to learn a control policy $\pi_{\theta}(\cdot)$ that enables an agent to maximize the expected total reward for given tasks. In this work, the policy is represented using a neural network, which is parameterized by $\theta$. An RL framework iterates between two key processes, policy evaluation and policy optimization. In policy evaluation, the robot dynamics is simulated; at each time step, an action is sampled from the policy and applied to control of the robot. Many excellent dynamics simulators nowadays are available to use, such as Pybullet, MuJoCo, Raisim, and IssacGym. Various algorithms are developed to optimize the policy, i.e., the network parameters $\theta$. Depending on the inclusion of a policy network, a value function network, or a combination of both, these algorithms could be one of the three categories, policy gradient, value function methods, or actor-critic algorithms. In this work, we use Proximal Policy Optimization (PPO) \cite{schulman2017proximal}, which is an actor-critic algorithm. Further, PPO is an off-policy algorithm and allows the same set of data to be used several times for policy updates.

\subsection{Nonlinear Model Predictive Control}
At the core of MPC is solving a sequence of finite-horizon trajectory optimization (TO) problems. In this work, we consider the case where the problem is non-linear and has multiple phases. The multi-phase TO problem is formulated as follows:
\begin{subequations}\label{eq_multiTO}
\begin{IEEEeqnarray}{cl}
\IEEEeqnarraymulticol{2}{l}{\min_{\u(\cdot)}  \sum_{i=1}^{n} \left[ \int_{t^+_{i-1}}^{t^-_i} l_i\big(\xi(t), \ui(t) \big) {\rm d}t + \Phi_i\big(\xi(t^-_i)\big) \right]}\\
    \text{subject~to} \ \ & \Dot{\x}^{[i]} = \f^{[i]}(\xi(t),\ui(t))\\
    & \gi\big(\xi(t), \ui(t)\big) \geq 0\\ 
     & \hi\big(\xi(t), \ui(t)\big) = 0,\\
     & \hi_e\big(\xi(t_i^-)\big) = 0,\\
     & \x^{[i+1]}(t_{i+1}^+) - \P_i(\xi(t_i^-)) = 0
\end{IEEEeqnarray}
\end{subequations}
where $i$ represents the phase index. In the context of legged locomotion, a phase refers to a period of time when the contact remains unchanged. The variables $\x$ and $\u$ denote the state and control respectively. The functions $l$ and $\Phi$ represent the running cost, and the terminal cost, respectively, $\f$ is the continuous dynamics, $\g$ the inequality constraint, $\h$ the equality constraint, $\h_e$ terminal constraint, and $\P$ the reset map.

Given a current state, MPC solves~\eqref{eq_multiTO} and applies the first control, moves the state to the next and resolves~\eqref{eq_multiTO}. The time between two MPC updates is often the same as the integration time step that is used to discretize~\eqref{eq_multiTO}. In more general settings, however, a greater update time could be used, implying that we could apply more than one control signal before updating MPC. In this work, we use this more general MPC setup. 

The formulation~\eqref{eq_multiTO} requires the contact schedule to be known. This requirement is helpful to solve multi-contact TO problems quickly and avoid the combinatorial complexity of contact sequence optimization. There are excellent works considering contact-implicit trajectory optimization such as \cite{posa2014direct, tassa2012synthesis, manchester2020variational}, but they are not sufficiently fast for online use with MPC. 

\section{Control Architecture Overview}
An overview of the proposed control architecture is illustrated in Fig~\ref{fig:overview}. We follow the work \cite{peng2020learning} to train RL policies that imitate motions of an animal dog. The RL policy is unrolled at 30 Hz to produce the state (CoM position, velocity, Euler angles, angular velocity, joint position) trajectory as well as the contact schedule. This information is then used to build the multi-phase TO problem \eqref{eq_multiTO} for MPC. The MPC simultaneously solves for the optimal foothold locations and GRFs. Swing trajectories are generated by interpolating the predicted foothold locations with a third-order Bezier curve, and a swing controller as in \cite{di2018dynamic} is employed to control the swing leg via a Cartesian impedance controller. The GRFs are converted to joint torques using $\J^{\top}\blambda$ for the control of the stance legs. The MPC runs at 50 Hz, whereas the main control loop and state estimation run at 500 Hz.

\section{Model Predictive Control with Hybrid Kinodynamic Model}
\label{sec_hkdmpc}
In this section, we discuss in detail the developed MPC controller that employs a hybrid kinodynamic (HKD) model and enables simultaneous optimization of the CoM trajectory, ground reaction forces, and foothold locations. A similar kinodynamic model was proposed in \cite{grandia2019feedback}, which considers leg kinematics in both swing and stance. In that work, the foothold locations are not modeled as part of the state and are implicitly optimized by modulating the joint angles to satisfy a non-slipping constraint during stance. This constraint is enforced in the velocity level and along the entire stance phase. As a result, it may take more iterations to solve the optimization problem. By contrast, our proposed HKD model considers a foot position constraint only at the instant of touchdown, resulting in an optimization problem that is easier to solve.

\subsection{Contribution: Hybrid Kinodynamic Model}
The HKD model considers the Single-Rigid-Body (SRBD) dynamics of the floating base, the leg kinematics for the swing legs, and a fixed foothold location for stance legs. The HKD model is formulated as follows:
\begin{subequations}
\begin{align}
        \Dot{\btheta} & = \T(\btheta)\bomega\label{eq_eulrate} \\
        \Dot{\p} &= \v \label{eq_comvel}\\
        \Dot{\bomega} &= \I^{-1} \Big(-\bomega\times\I\bomega + \R^{\top}_B\sum_{j=1}^4 s_j(\p_{f_j} - \p) \times\blambda_{f_j} \Big)\label{eq_ang_acc} \\
        \Dot{\v} &= \g + \frac{1}{m}\sum_{j=1}^4 s_j \blambda_{f_j}\label{eq_acc} \\
        \Dot{\p}_{f_j} & = 0 \quad \ \  \text{if $j$ in stance} \label{eq_stance}\\
        \Dot{\q}_j &= \u_J \quad \text{if $j$ in swing}\label{eq_swing}
\end{align}
\end{subequations}
where $\btheta$ denotes the Euler angles, $\p$ the Center of Mass (CoM) position of the body, $\bomega$ and $\v$ respectively are angular and linear velocities of the body, $\R_B$ is the rotation matrix from the world frame to the body frame, $\I$ is the rotational inertia relative to the body frame, $m$ is the body mass, $\g$ is the gravity vector in world frame, $\p_f$ and $\blambda_f$ are the foothold location and ground reaction force (GRF) respectively, $s \in \{0,1\}$ is the contact indicator, $\q$ is the joint angle, $\u_J$ is the commanded joint velocity, and $j \in \{1,2,3,4\}$ denotes the leg index. The matrix $\T$ transforms the angular velocity to the rate of change of Euler angles. The variables $\bomega$ and $\I$ are expressed in the body frame, whereas $\p$, $\v$, $\p_f$, and $\blambda_f$ are in the world frame.

The Eqs.~\eqref{eq_eulrate}-\eqref{eq_acc} represent SRBD dynamics, whereas the Eqs.~\eqref{eq_stance}-\eqref{eq_swing} constrain the foothold locations and the joint angles at the kinematics level, and are complementary to each other. When a leg is in swing, its joint angles are modulated for reference tracking, whereas when in stance, its foothold location is fixed and the joint angles are ignored. 


\subsection{Reset Map}
Contacts are frequently established and broken during quadruped locomotion. Therefore, reset maps need to be specified for transitioning from swing to stance and vice versa. The foothold locations are computed using the leg forward kinematics at the end of the swing and held constant thereafter. The ab/ad, hip, and knee joint angles are set to default values ($[0,-0.8,1.8]$) in radian at the beginning of a swing, and follow Eq.~\eqref{eq_swing} afterwards. The reset map for the $j^{\text{th}}$ foot is summarized as follows:
\begin{subequations}\label{eq_resetmap}
\begin{align}
    \p_{f_j}^+ &= \text{FW}_j(\q_j^-) \hspace{20pt} \text{swing}\rightarrow\text{stance} \label{eq_resetmap_swing}\\
    \q_j^+ &= [0, -0.8, 1.8] \hspace{7pt} \text{stance}\rightarrow\text{swing} \label{eq_resetmap_stance}
\end{align}
\end{subequations}
where $\text{FW}(\cdot)$ represents the forward kinematics, the superscript `-' denotes the end state of the previous phase, and `+' the start of the next phase.
\subsection{Constraints}
Given the reset map~\eqref{eq_resetmap_swing}, it is important to ensure the computed foothold locations are on the ground a touchdown. Therefore, an equality terminal constraint is enforced at the end of swing to set the vertical ($z$) component of the foot position:
\begin{equation}
    \begin{bmatrix}
    0 & 0 & 1
    \end{bmatrix} \text{FW}_j(\q_j^-) = 0.
\end{equation}
Two types of inequality constraints are employed in this work. To prevent slipping and avoid negative normal GRF a linear approximation of the friction cone constraint is enforced during stance:
\begin{equation}
    f_z \geq 0, \abs{f_x } \leq \mu f_z, \abs{f_y} \leq \mu f_z
\end{equation}
To prevent swing leg collisions with the ground, foot positions are constrained to be above the ground  during swing:
\begin{equation}\label{eq_tdconstr}
    \begin{bmatrix}
    0 & 0 & 1
    \end{bmatrix} \text{FW}_j(\q_j^-) > 0 \ \forall j \text{ in swing}
\end{equation}
\subsection{Cost Function}
The cost function employed in this work consists of a tracking cost and two regularization terms. The tracking cost is a quadratic function that penalizes deviations from the roll-out trajectory. For each phase $i$ where contacts remain unchanged, the running cost function is defined as
\begin{multline}
    l_{\text{track}} = \int_{t^+_{i-1}}^{t^-_i} 
    \snorm{[\delta\btheta^{\top} \ \delta\p^{\top} \  \delta\bomega^{\top} \delta\v^{\top}]}_{\Q_b} + \\ \snorm{\Bar{\S}\delta\q_J}_{\Q_J} + \snorm{\S\blambda}_{\R_{\lambda}} dt
\end{multline}
where $\delta \cdot$ represents the deviation, and $\Q_b$, $\Q_J$ and $\R_{\lambda}$ are positive definite weighting matrices. The matrix $\S$ is a diagonal matrix concatenating the contact status of each foot whereas $\Bar{\S}$ concatenates the swing status. The terminal cost is defined similarly but without the last term.

\HL{A foot regularization term is used to encourage the optimized foot placements towards reference positions, and is defined as follows
\begin{equation}
    l_{\text{foot}} = \sum_j^4 \int_{t^+_{i-1}}^{t^-_i} \snorm{\p_{f_j} - \p - \p_{\text{rel},j}} dt
\end{equation}
where $\p_{\text{rel},j}$ is the foot position relative to the CoM position for the $j^{\text{th}}$ foot, expressed in the world frame. This regularization is helpful to prevent self collision, especially for the pacing motion, where the optimal foot placements are approximately on the sagittal plane if the regularization is not used. In addition, a smoothness regularization term is used to encourage the solution to stay close to the previous solution. }

\begin{figure*}[t]
    \centering
    \includegraphics[width=0.8\linewidth]{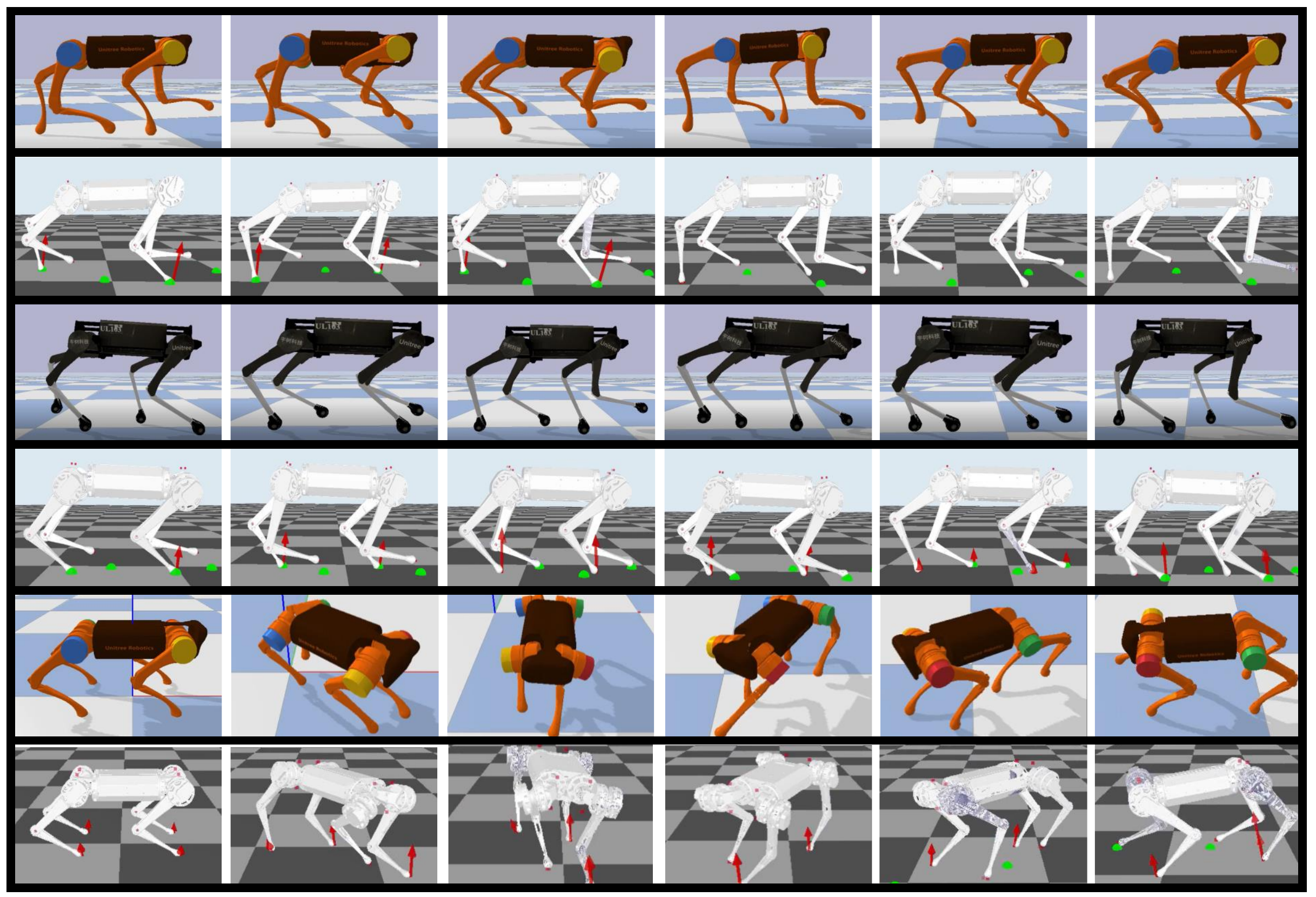}
    \caption{Time-series snapshots of motions using original RL policies and the retargeted motion using the proposed framework. Top row: A1 trotting using the RL policy. Second row: Retargeted trotting motion on Mini Cheetah using the proposed framework. Third row: Laikago pacing using the RL policy. Bottom row: Retargeted pacing motion on Mini Cheetah using the proposed framework.}
    \label{fig:snapshots_ofall}
\end{figure*}

\begin{figure}[t]
    \centering
    \includegraphics[width = 0.85\linewidth]{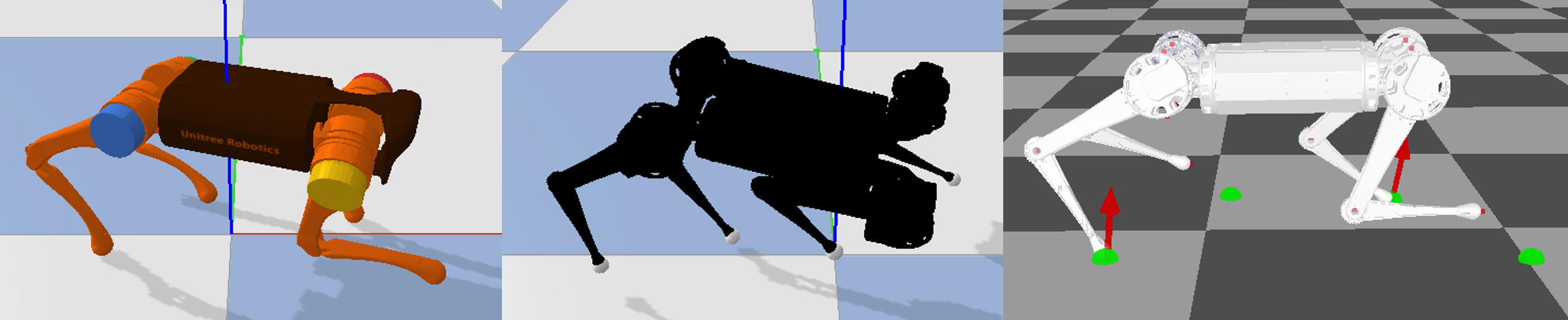}
    \caption{Left:A1 robot executes the RL policy in PyBullet. Middle: Mini Cheetah executes the RL policy in PyBullet. Right: Policy retargetting on Mini Cheetah in the Mini Cheetah Simulator.}
    \label{fig:performance_comp}
\end{figure}

\section{Simulation Results}\label{sec_sim}
The proposed framework is tested in simulation on retargeting of three locomotion policies, a trotting policy, a pacing policy, and a turning policy. The locomotion policies are originally trained for the Unitree A1 and Laikago robots, respectively, and the retargeted machine is the MIT Mini Cheetah robot \cite{katz2019mini}. These three robots differ in geometry and weights. The Mini Cheetah weighs 9 kg, and stands around 0.3 m tall. The A1 robot has similar size, but weighs 12.7 kg. The Laikago robot is of much larger size, weighs 22 kg, and stands around 0.6 m tall. The simulation results demonstrate successful motion retargeting in zero shot despite the geometry and weight differences.

\subsection{Implementation Details}

We follow the work in \cite{peng2020learning} to train the three locomotion policies aforementioned to imitate quadruped animal motions. The policies are trained in the physics engine PyBullet \cite{coumans2021}, using PPO \cite{schulman2017proximal}. Each training collects around 200 million samples, and takes around six days on a super computer using a 8-core CPU in parallel. We refer the readers to \cite{peng2020learning} for the implementation details of policy training. 

\HL{The RL locomotion policies are unrolled offline at 30 Hz to generate the reference trajectory. The foot contacts and the reference foot placements are acquired by performing collision detection in PyBullet. We observe that the robot feet may sometimes bounce at touchdown, resulting in frequent switching between stance and swing. A debouncing scheme is employed to address this problem, which averages the contact status over the subsequent five time steps once a contact is detected. The contact is considered active if the average is above 0.5.}

The HKD-MPC discussed in Section~\ref{sec_hkdmpc} is solved using a customized Hybrid System Differential Dynamic Programming (HSDDP) solver \cite{li2020hybrid}, where we use Forward Euler integration with time step 0.011 s, and planning horizon of 0.462 s. The HKD-MPC re-plans at 50 Hz, and each re-planning is warm-started with the solution from the previous plan. We found that it is sufficient to run DDP up to three iterations without losing the control performance, which takes on average 6 ms on the development computer (a ThinkPad Laptop with a 8-core, 2.5GHz, 11th-Gen Intel CPU). The the main control loop and the state estimator \cite{bledt2018cheetah} run at 500 Hz and communicate with the MPC module via LCM \cite{5649358}. \HL{The optimization introduces a 6 ms policy lag between the time a MPC update request is sent and the time the optimal solution is received. To account for this policy lag, we use the MPC control signals that are the closest to the current time, and ignore those earlier. Communication lags are not considered as we do not find any issue with the current implementation.} The cost functions employed for HKD-MPC could be found in our code. \footnote{\url{https://github.com/heli-sudoo/HKDMPC.git}}

\subsection{Simulation Results}

The locomotion RL policies are executed in PyBullet, and the retargeted motions with HKD-MPC are assessed in a high-fidelity dynamics simulator designed for the Mini Cheetah. Due to space limitation, only part of these results are presented here and summarized in Fig.~\ref{fig:snapshots_ofall}, i.e., A1 trotting policy, Laikago pacing policy, A1 turning policy, and the resulting retargeted motions. The readers are encouraged to check the accompanying video for the complete set of results. With the proposed control architecture in Fig.~\ref{fig:overview}, we could successfully clone the behaviours of RL policies trained for one robot onto another robot, despite large difference in their dimensions, the inertial parameters, and the internal dynamic attributes. The success is accounted for by the way the policy roll-out is handled in the proposed control architecture: the roll-out is used as a reference for MPC, which synthesizes a motion that is dynamically-feasible to the targeted platform while tracking the roll-out as well. By comparison, an immediate idea is to directly execute the RL policies on the retarget platform, given that training a new policy may take another couple of days. Fig.~\ref{fig:performance_comp} shows the result of directly executing the A1 trotting policy on the Mini Cheetah--Mini Cheetah falls down in just a couple of steps. This failure is a consequence of the sensitivity of RL policies to model mismatch, which also accounts for the well-known sim-to-real transfer problem. Dynamics randomization is one way of addressing this problem, but it requires additional training, and it cannot account for large model mismatch such as from Laikago to MC. From this perspective, the proposed control architecture not only provides a solution on retargeting motions among different platforms without further training, but also offer an opportunity to mitigate the sim-to-real problem of RL.

\begin{figure}[t]
    \centering
    \includegraphics[width = .8\linewidth]{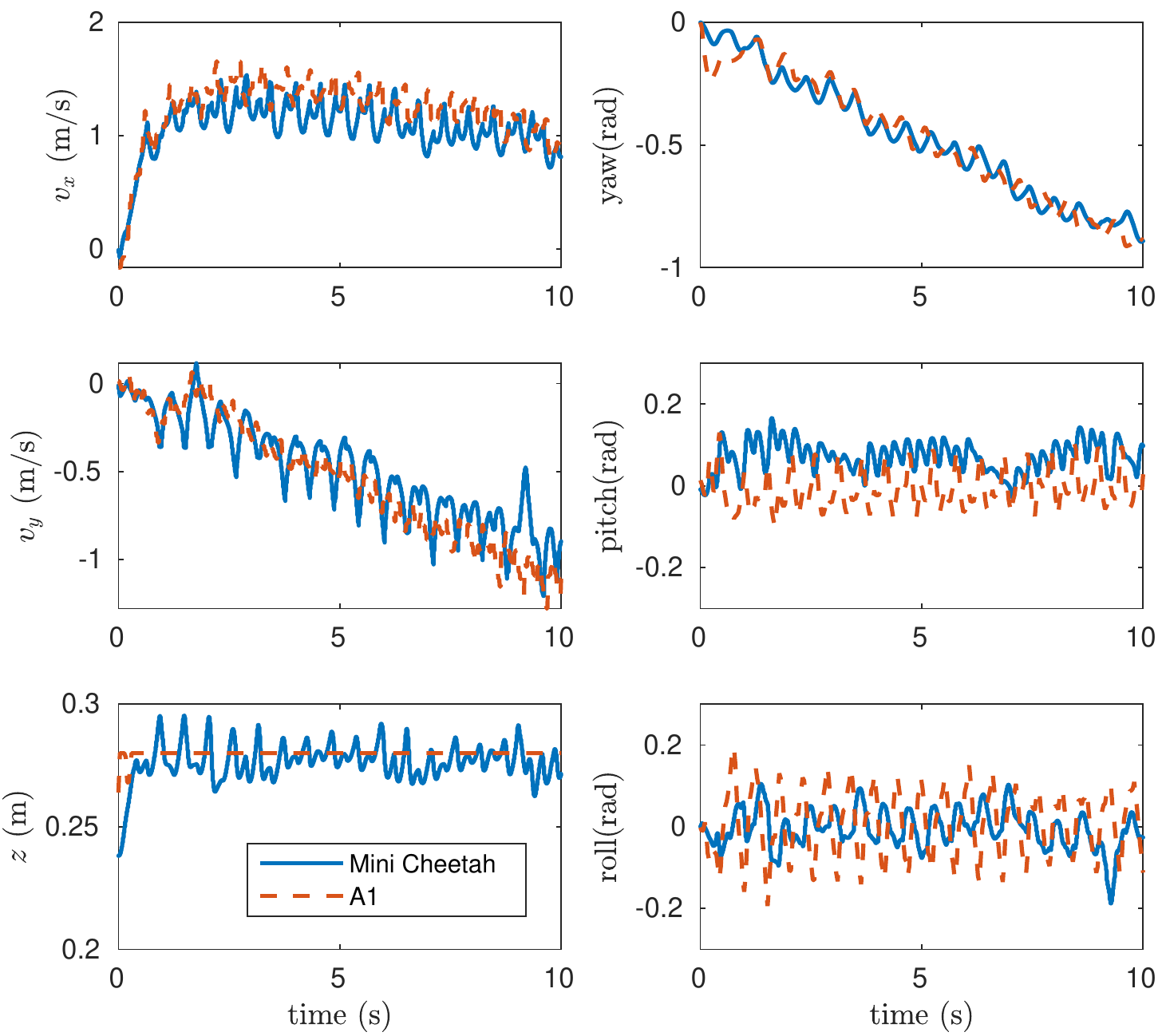}
    \caption{Retargeted trotting motion on the Mini Cheetah (solid curve) versus the RL trotting policy roll-out on the A1 robot (dashed curve).}
    \label{fig:tracking_trot}
\end{figure}

\begin{figure}[t]
    \centering
    \includegraphics[width = 0.8\linewidth]{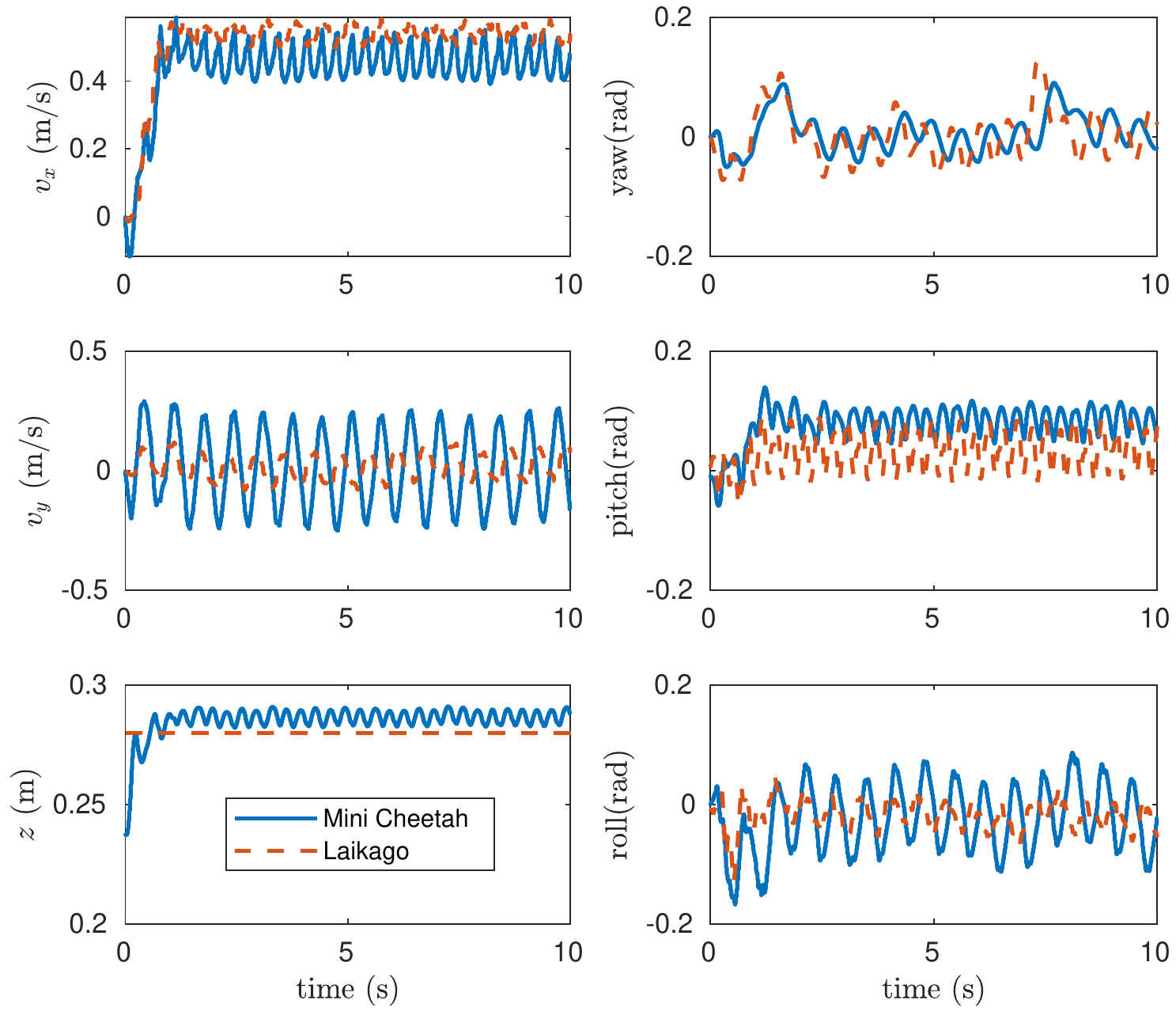}
    \caption{Retargeted pacing motion on the Mini Cheetah (solid curve) versus the RL pacing policy roll-out on Laikago (dashed curve). The CoM velocities of the policy roll-out are scaled by half, and the CoM height is truncated at 0.28 m.}
    \label{fig:tracking_pace}
\end{figure}

\begin{figure}[t]
    \centering
    \includegraphics[width = 0.8\linewidth]{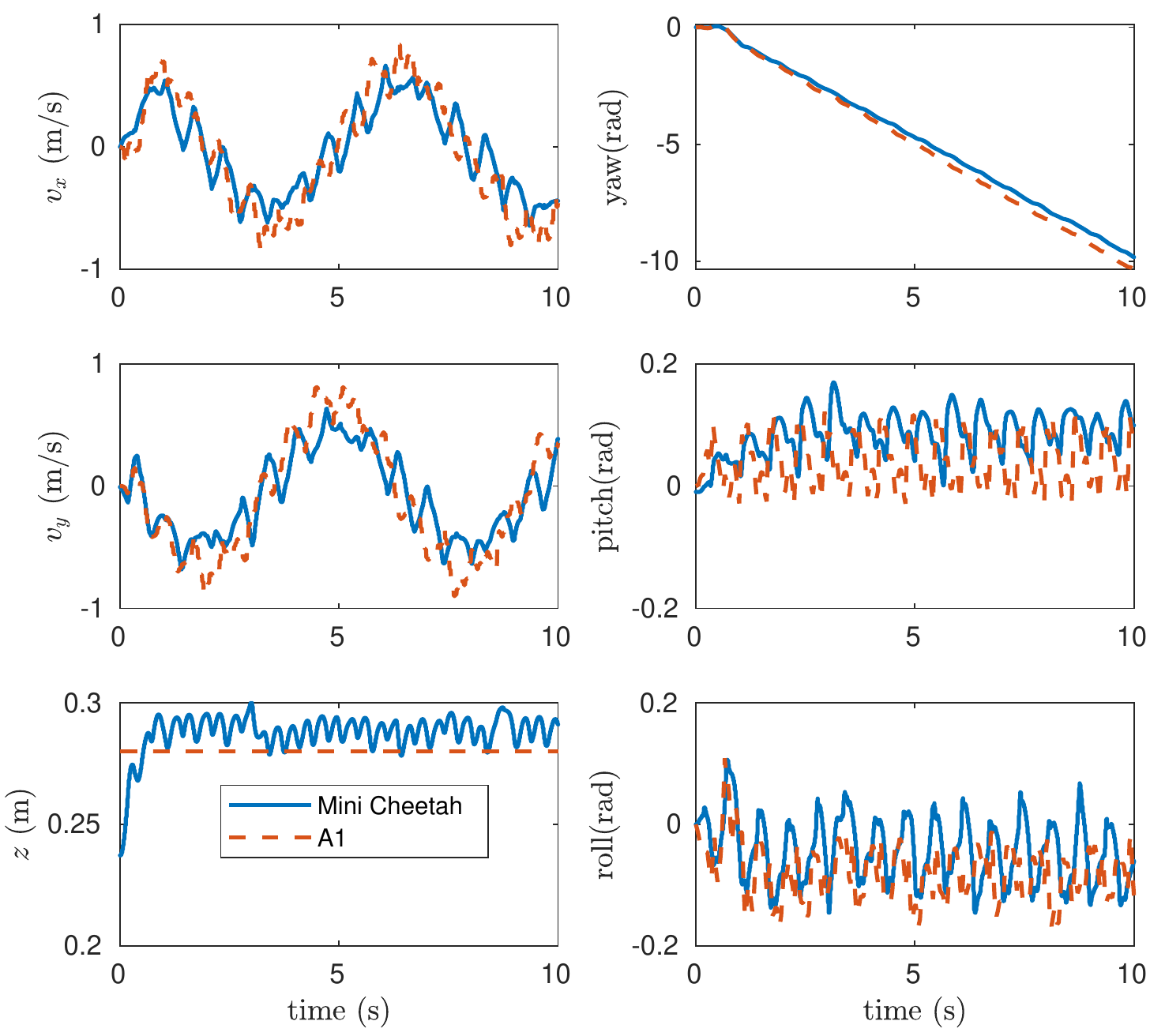}
    \caption{Retargeted turning motion on the Mini Cheetah (solid curve) versus the RL turning policy roll-out on A1 (dashed curve).}
    \label{fig:tracking_turn}
\end{figure}

\HL{The detailed retargeting performance of the proposed control architecture is present in Fig.~\ref{fig:tracking_trot}, Fig.~\ref{fig:tracking_pace}, and Fig.~\ref{fig:tracking_turn} for trotting, pacing, and turning respectively. The RL policy roll-outs and the retargeted motions are compared in terms of the fore/aft and lateral CoM velocities $v_x$, $v_y$, CoM height $z$, and the roll-pitch-yaw angles. The retargeted motions and the RL policy roll-out are in general reasonably close, demonstrating good retargeting performance. However, we do note that the retargeted motions have larger variations, especially for forward and lateral velocities. Several reasons could be accounted. First, the swing leg controller is purely a PD controller, and has high stiffness at touchdown. Employing an impedance controller would allow smaller stiffness and may help mitigate the problem. Second, the foot regularization introduces extra angular momentum. More careful tuning of foot regularization and tracking cost could be helpful. Note that since the pacing policy is trained for Laikago which has larger size, the CoM velocities are scaled by half, and the robot height is truncated to 0.28 m. We find this simple strategy to work reasonably well for our problems. A more general strategy such as an an-isotropic scaling would be needed to handle a larger variety of robot morphologies.}


\begin{figure*}[t]
    \centering
    \includegraphics[width=0.8\textwidth]{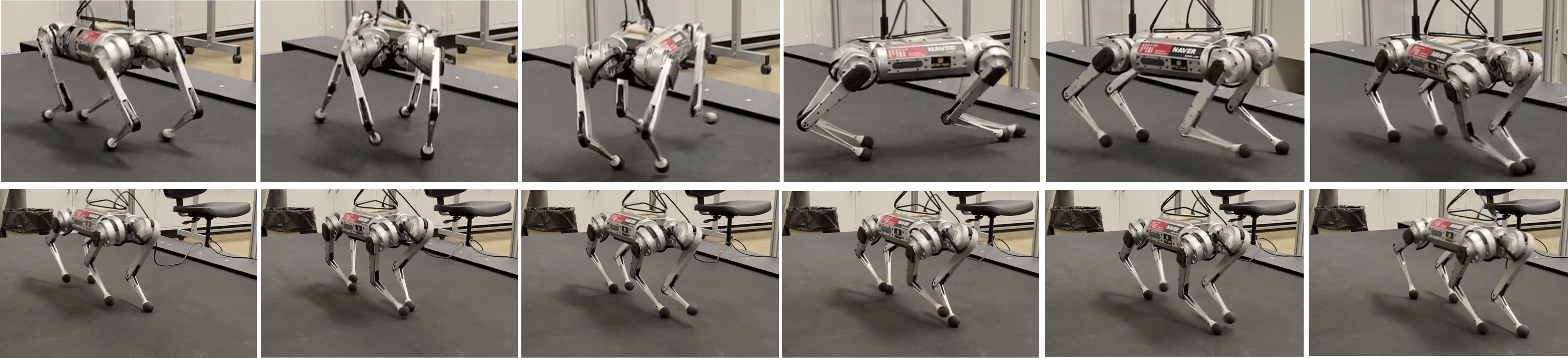}
    \caption{Time-series snapshots of the retargeted motions on the Mini Cheetah robot hardware. Top: turning motion. Bottom: pacing motion.}
    \label{fig:hardware}
\end{figure*}

\section{\HL{Hardware Results}}\label{subsec_exp_results}
We test the performance of the proposed framework on the Mini Cheetah hardware to retarget the three RL locomotion policies that are trained for A1 and Laikago, as discussed in Section~\ref{sec_sim}. The implementation details remain the same as in simulation except that the main control loop and the state estimator now run on the robot computer. 

\subsection{Hardware Test Results}
All the three locomotion policies are successfully retargeted to the robot hardware using the proposed pipeline. Figure~\ref{fig:hardware} depicts time-series snapshots of turning and pacing motions. The readers are encouraged to check the accompanying video for the complete sets of the experiments including a trotting motion.

 A post-processing step was made for the retargeting of the turning motion. Note in Fig.~\ref{fig:tracking_turn} that there is a fast yaw change taking place at about 1 second. We were not able to retarget this motion on the robot hardware due to the collision of the front two legs when leg crossing occurs. The failure is not observed in simulation as self-collision was not turned on. To get around this problem, we deliberately skip the first circle of the turning motion and retarget the remaining motion. However, we do note that a better solution is to consider the self-collision in MPC, which is under development. A closer examination of the RL turning policy reveals that it is the sliding contact that results in the fast yaw change. Handling sliding contact is in general hard and is an open problem in MPC.

\section{\HL{Discussions}}
Previous sections have shown that the proposed framework can retarget a variety of RL locomotion policies trained for one robot to the other robots of different geometry and weights. The proposed framework explores a novel method of combining RL and MPC. In this section, we discuss the limitations of the proposed framework, and the benefits of combining the two in comparison pure RL and MPC.

One limitation of the proposed framework is that it cannot handle sliding contacts. Contacts are assumed stationary in the HKD-MPC. If sliding contacts present in the RL policy roll-out, chances are that the robot would fall. The fall may result from self-collision as discussed in Section~\ref{subsec_exp_results}, or from the solver failing to find a feasible solution. One solution is to design a proper reward that penalizes foot slip when training the RL policy, but this approach may not be helpful on slippery ground such as ice.

The other limitation is that the swing controller is not collision-free. The current method optimizes foot placements and interpolates a cubic polynomial for swing foot, which is not obstacle-aware. Future work would investigate adding obstacle clearance constraints to the HKD-MPC, and a swing controller to track the collision-free joint trajectory.

The proposed framework essentially uses RL for trajectory generation and MPC as a tracking controller. The motivation for this structure is that if we unroll the RL policy online, the RL policy has the potential of discovering new contacts for MPC in response to external disturbances and new terrains. This feature could make the proposed framework more advantageous than conventional MPC alone, which assumes a fixed contact sequence. Further, pure RL methods are subject to sim-to-real transfer problems. Given that the proposed framework successfully retargets RL locomotion policies to robots of significantly different size, it has the potential of attacking the sim-to-real transfer problems resulting from model mismatch. 

\section{Conclusions}
In this work, we demonstrate that motor skills could be transferred in zero shot between robots of different dimensions and weights. We do so by generating reference robot motions and their contact schedules using RL policies, while retargeting the motion to another robot using MPC. 
One important reason is that the MPC can synthesize motions that are feasible for the target robot while still tracking the reference motions. Although the reference motions and  accompanying contact schedule are not necessarily dynamically-feasible for the target robot, they provide a good seed for MPC after simple adjustments. 
The proposed method avoids the need to retrain a policy on the target robot to achieve similar behaviours, and simplifies the process of manually crafting reference motions for complex behaviours.

\bibliographystyle{IEEEtran}
\bibliography{ms.bib}

\end{document}